\documentclass[journal,twoside,web]{IEEEtran}
\usepackage{tmi}
\usepackage{cite}
\usepackage{amsmath,amssymb,amsfonts}
\usepackage{algorithmic}
\usepackage{graphicx}
\usepackage{textcomp}

\usepackage{amsmath}
\usepackage{multirow}
\usepackage{multicol}
\usepackage{makecell}
\usepackage[ruled]{algorithm2e}
\usepackage{subfigure}
\usepackage{amsmath}
\usepackage[utf8]{inputenc} 
\usepackage[T1]{fontenc}    
\usepackage{url}            
\usepackage{booktabs}       
\usepackage{amsfonts}       
\usepackage{nicefrac}       
\usepackage{microtype}      
\usepackage{xcolor}         
\makeatletter
\newcommand*\bigcdot{\mathpalette\bigcdot@{.5}}
\newcommand*\bigcdot@[2]{\mathbin{\vcenter{\hbox{\scalebox{#2}{$\m@th#1\bullet$}}}}}
\makeatother

\usepackage{tikz,xcolor,hyperref}
\definecolor{Revision2}{RGB}{0, 0, 0}
\definecolor{Revision3}{RGB}{0, 0, 0}
\definecolor{Revision}{RGB}{0, 0, 0}
\definecolor{lime}{HTML}{A6CE39}
\DeclareRobustCommand{\orcidicon}{
\begin{tikzpicture}
\draw[lime, fill=lime] (0,0)
circle[radius=0.16]
node[white]{{\fontfamily{qag}\selectfont \tiny \.{I}D}}; 
\end{tikzpicture}
\hspace{-2mm}
}
\foreach \x in {A, ..., Z}{%
\expandafter\xdef\csname orcid\x\endcsname{\noexpand\href{https://orcid.org/\csname orcidauthor\x\endcsname}{\noexpand\orcidicon}}
}

\def\BibTeX{{\rm B\kern-.05em{\sc i\kern-.025em b}\kern-.08em
    T\kern-.1667em\lower.7ex\hbox{E}\kern-.125emX}}
\begin{document}

\title{ScribFormer: Transformer Makes CNN Work Better for Scribble-based Medical Image Segmentation}

\author{Zihan Li, Yuan Zheng, Dandan Shan, 
        Shuzhou Yang, Qingde Li, Beizhan Wang, \\Yuanting Zhang,~\IEEEmembership{Fellow,~IEEE}, Qingqi Hong,~\IEEEmembership{Member,~IEEE}, Dinggang Shen,~\IEEEmembership{Fellow,~IEEE}
\vspace{-6mm}  
\thanks{This work was supported in part by National Natural Science Foundation of China (grant numbers 62131015, 62250710165, U23A20295), Science and Technology Commission of Shanghai Municipality (STCSM) (grant numbers 21010502600), and The Key R\&D Program of Guangdong Province, China (grant numbers 2021B0101420006).}%
\thanks{Zihan Li is with Xiamen University and the Department of Bioengineering, University of Washington, Seattle, WA 98195, USA.}
\thanks{Yuan Zheng, Dandan Shan and Beizhan Wang are with Xiamen University, Xiamen 361005, China.}
\thanks{Shuzhou Yang is with Peking University, Shenzhen, 518055, China.}
\thanks{Qingde Li is with University of Hull, Hull, HU6 7RX, UK.}
\thanks{Yuanting Zhang is with the Department of Electronic Engineering at the Chinese University of Hong Kong, Shatin, Hong Kong, China, and also Hong Kong Institute of Medical Engineering, Taipo, Hong Kong, China.}
\thanks{Qingqi Hong is with Xiamen University, Xiamen 361005, China, and also Hong Kong Centre for Cerebro-cardiovascular Health Engineering, Hong Kong, China. (e-mail: hongqq@xmu.edu.cn).}
\thanks{Dinggang Shen is with the School of Biomedical Engineering \& State Key Laboratory of Advanced Medical Materials and Devices, ShanghaiTech University, Shanghai 201210, China, Shanghai United Imaging Intelligence Co., Ltd., Shanghai 200230, China, and also Shanghai Clinical Research and Trial Center, Shanghai, 201210, China. (e-mail: dinggang.shen@gmail.com).}
\thanks{Zihan Li and Yuan Zheng have equal contribution to this work. }
\thanks{Corresponding authors: Qingqi Hong and Dinggang Shen.}
}

\markboth{ IEEE TRANSACTIONS ON MEDICAL IMAGING, VOL. XX, NO. XX, XXXX 2023}%
{Li \MakeLowercase{\textit{et al.}}: ScribFormer: Transformer Makes CNN Work Better for Scribble-based Medical Image Segmentation}


\maketitle

\begin{abstract}
Most recent scribble-supervised segmentation methods commonly adopt a CNN framework with an encoder-decoder architecture. Despite its multiple benefits, this framework generally can only capture small-range feature dependency for the convolutional layer with the local receptive field, which makes it difficult to learn \textcolor{Revision2}{global shape information} from the limited information provided by scribble annotations. To address this issue, this paper proposes a new CNN-Transformer hybrid solution for scribble-supervised medical image segmentation called ScribFormer. The proposed ScribFormer model has a triple-branch structure, i.e., the hybrid of a CNN branch, a Transformer branch, and an attention-guided class activation map (ACAM) branch. Specifically, the CNN branch collaborates with the Transformer branch to fuse the local features learned from CNN with the global representations obtained from Transformer, which can effectively overcome limitations of existing scribble-supervised segmentation methods. Furthermore, the ACAM branch assists in unifying the shallow convolution features and the deep convolution features to improve model's performance further. Extensive experiments on two public datasets and one private dataset show that our ScribFormer has superior performance over the state-of-the-art scribble-supervised segmentation methods, and achieves even better results than the fully-supervised segmentation methods. The code is released at {\href{https://github.com/HUANGLIZI/ScribFormer}{https://github.com/HUANGLIZI/ScribFormer}}.
\end{abstract}

\vspace{-2mm}
\begin{IEEEkeywords}
Transformer, Medical image segmentation, Scribble-supervised learning.
\end{IEEEkeywords}

\section{Introduction}
\vspace{-2mm}
\begin{figure}[ht]
  \centering
  \includegraphics[width=\linewidth]{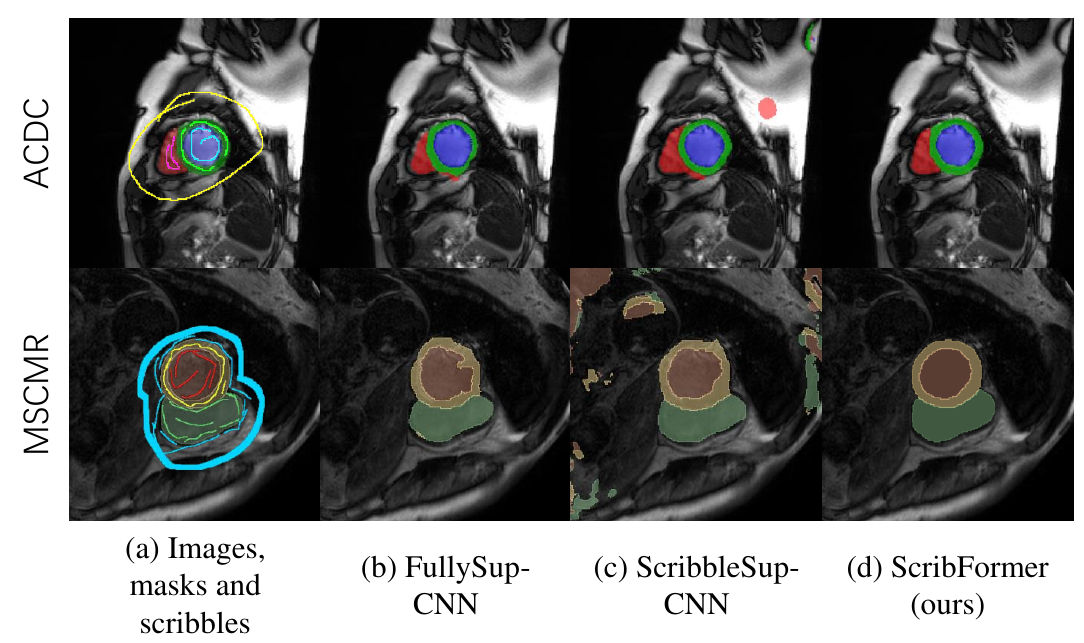}
  \vspace{-6mm}
  \caption{
  Performance comparison of segmentation results across various methods: (a) Input images, masks, and scribble annotations, (b) CNN-based fully-supervised method (UNet++ \cite{unet++}), (c) CNN-based scribble-supervised method (UNet++), and (d) Our proposed ScribFormer.
  }
  \label{comparison}
  \setlength{\belowcaptionskip}{-1cm}
  \vspace{-3mm}
\end{figure}
\IEEEPARstart{D}{eep} convolutional neural networks (CNN) have produced highly promising results in the automatic segmentation of medical images.
However, their advancement is hindered by the lack of sufficiently large and fully labeled training datasets. 
Generally, most deep CNN methods require large-scale images with precise, dense, pixel-level annotations for model training. 
Unfortunately, manual annotation of medical images is a time-consuming and expensive process that requires skilled clinical professionals. To address this challenge, recent researchers have been developing novel techniques that do not rely on fully and accurately labeled datasets. One such technique is weakly-supervised learning, which trains a model using loosely-labeled annotations such as points, scribbles, and bounding boxes for areas of interest. These approaches aim to reduce burden on clinical professionals while still achieving high-quality segmentation results.
Compared to other annotation methods, such as bounding boxes and points, scribble-based learning (where masks are provided in the form of scribbles) offers greater convenience and versatility for annotating complex objects in images \cite{tajbakhsh2020embracing}.

Existing CNN-based scribble learning models can be broadly classified into two categories according to the ways of using the limited information provided by scribble annotations. 
The first category focuses on learning adversarial \textcolor{Revision2}{global shape information} with a conditional mask generator and a discriminator \cite{larrazabal2020post, MAAG, zhang2020accl}, which generally requires \textbf{extra fully-annotated masks}.
The second category, on the other hand, utilizes targeted training strategies or elaborated structures directly on the scribbles \cite{lin2016scribblesup, kim2019mumford, EM}. However, the process of scribble-supervised training may generate \textbf{noisy labels} that can degrade segmentation performance of trained models. As shown in Fig. \ref{comparison}, compared to the fully-supervised CNN (b), the scribble-supervised CNN (c) trained only on a few labeled pixels may lead to extra segmentation areas with noise.
In recent years, several studies have attempted to expand scribble annotation by leveraging data enhancement strategies \cite{Zhang_2022_CycleMix} or generating pseudo labels \cite{luo2022scribble} to address the issue of noisy labels. Nevertheless, the principal obstacle of scribble-based segmentation still lies in training a segmentation model with inadequate supervision information, as a scribble is an inaccurate representation for the area of interest.

Our work delves into the use of scribble annotations to efficiently train high-performance medical image segmentation models.
To address the \textbf{first issue} of learning \textcolor{Revision2}{global shape information} without the availability of fully-annotated masks, we investigate the utilization of Transformers \cite{transformer} for weakly-supervised semantic segmentation (WSSS). Generally, the Vision Transformer (ViT) \cite{dosovitskiy2020image} leverages multi-head self-attention and multi-layer perceptrons to capture long-range semantic correlations, \textcolor{Revision}{ which are
 crucial for both localizing entire objects and implicitly learning \textcolor{Revision2}{global shape information} through subsequent ACAM branches.}
However, in contrast to CNN, ViT often ignores local feature details of objects that are also important for WSSS applications.
Hybrid combinations of CNN and ViT architectures have been developed  \cite{wu2021cvt, peng2021conformer, shan2022c2fvl, li2023scribblevc} to take advantage of their respective strengths. In particular, we utilize a CNN branch and a Transformer branch to fuse local features and global representations interdependently at multiple scales, which can achieve superior performance on the segmentation task.

To address \textbf{the second issue} of expanding scribble annotations for WSSS, class activation maps (CAMs) \cite{zhou2016cam,li2022semi} are often used to generate initial seeds for localization. However, the pseudo labels generated from CAMs for training a WSSS model have an issue of partial activation, which generally tends to highlight the most discriminative part of an object instead of the entire object area \cite{wang2020self, lee2021anti}. Recent work \cite{peng2021conformer} has pointed out that the reason may be the intrinsic characteristic of CNNs, i.e., the local receptive field only captures small-range feature dependencies. Although various methods have been proposed to identify an activation area aligned with the entire object region \cite{wang2020self,lee2021anti}, little work has directly addressed the local receptive field deficiencies of the CNN when applied to WSSS. Motivated by these observations, we incorporate an attention-guided class activation map (ACAM) branch into the network. In the ACAM branch, instead of implementing traditional CAMs that generally only highlight the most discriminative part, 
\textcolor{Revision}{attention-guided CAMs restore activation regions missed in various encoding layers during the encoding process. This approach can achieve the reactivation of mixed features and focuses on the whole object.}
Moreover, ACAMs-consistency is employed to penalize inconsistent feature maps from different convolution layers, in which the low-level ACAMs are regularized by the high-level ACAM generated from feature of last CNN-Transformer branch layer. 

In this paper, we propose a novel weakly-supervised model for scribble-supervised medical image segmentation, named ScribFormer, which consists of a triple-branch network, i.e., the hybrid CNN and Transformer branches, along with an attention-guided class activation map (ACAM) branch. Specifically, in the hybrid CNN and Transformer branches, the global representations and the local features are mixed to enhance each other.
Fig. \ref{comparison} shows two examples of segmentation results generated by different models. It can be observed that the famous CNN-based UNet model could fail in the scribble supervision-based segmentation, which generates several invalid prediction results in background regions (Fig. \ref{comparison} (c)). On the contrary, our ScribFormer model can overcome this problem and generate much more satisfactory results (Fig. \ref{comparison} (d)) based on the proposed triple-branch architecture. The hybrid architecture can leverage detailed high-resolution spatial information from CNN features and also the global context encoded by Transformers, which is of great help for scribble-supervised medical image segmentation. 

The contributions of this paper are summarized as follows:

\begin{itemize}
\item To the best of our knowledge, our ScribFormer is the first Transformer-based solution for scribble-supervised medical image segmentation, which employs a hybrid CNN-Transformer architecture to leverage both the local detailed high-resolution spatial information learned from CNN features and the global context encoded by Transformers.

\item In ScribFormer, Transformers have emerged as the architecture with the \textbf{innate} global self-attention mechanism, which can reduce invalid prediction results in background regions. Meanwhile, the global representation captured by Transformers implicitly refines the ACAMs generated from the CNN branch, which can address the partial activation issue of CAMs caused by the inherent deficiencies of CNN’s local receptive field.

\item We propose the ACAMs-consistency loss to train the low-level convolutional layers under the supervision of high-level convolutional features, which can further improve model's performance. ScribFormer has been evaluated on \textcolor{Revision}{three} datasets, i.e., ACDC, MSCMR\textcolor{Revision}{, and HeartUII}, and achieved superior segmentation performance over state-of-the-art scribble-supervised methods.
\end{itemize}

\begin{figure*}[htbp]
\centering 
\setlength{\abovecaptionskip}{-1mm}
\includegraphics[width=0.95\textwidth]{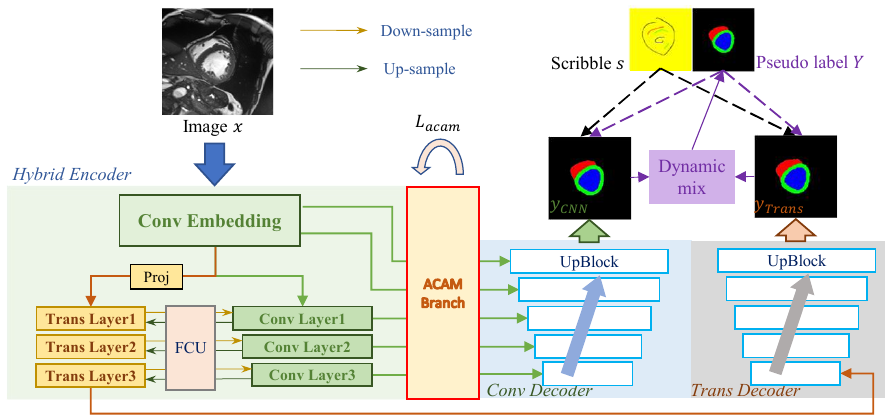}
\caption{\textcolor{Revision}{Overview of our proposed ScribFormer. The framework consists of the hybrid CNN-Transformer encoders, the CNN decoder, the Transformer decoder, and the attention-guided class activation map (ACAM)  branch. Both the CNN prediction $y_{CNN}$ and the Transformer prediction $y_{Trans}$ are compared separately with the scribble annotations and the dynamically mixed pseudo label. Furthermore, the ACAM branch compares multi-scale ACAMs with ACAM-consistency.}}
\label{framework}
\vspace{-6mm}
\setlength{\belowcaptionskip}{-1cm}
\end{figure*}
\vspace{-3mm}
\section{Related Works}
\subsection{Transfomers for Medical Image Segmentation}

Medical image segmentation plays a crucial role in many fields, such as brain segmentation \cite{jia2012iterative,wang2011automatic}, registration \cite{fan2018adversarial}, and disease diagnosis \cite{ fan2007multivariate}. 
A new paradigm for medical image segmentation has evolved thanks to the success of Vision Transformer (ViT) \cite{dosovitskiy2020image} in many computer vision fields. 
Generally, the Transformer-based models for medical image segmentation can be classified into two types: 1) ViT as the main encoder and 2) ViT as an additional encoder \cite{li2022transforming}.
In the first type, the global attention-based ViT is utilized as the main encoder and connected to the CNNs-based decoder modules, such as the works presented in \cite{swinunetr,qiu2023corsegrec,li2022lvit,Hatamizadeh_2022_WACV,li2022tfcns,li2023nnsam}.
The second model type utilizes Transformers as the secondary encoder after the main encoder CNN. There are several representative works following this widely-adopted structure, including TransUNet \cite{chen2021transunet}, TransUNet++ \cite{wang2022multiscale}, CoTr \cite{xie2021cotr}, SegTrans \cite{li2021medical}, TransBTS \cite{wang2021transbts}, and so on.
In the hybrid models, ViT and CNN encoders are combined to take the medical image as input, and then the embedded features are fused to connect to the decoder. This multi-branch structure provides the benefits of simultaneously learning global and local information, which has been utilized in several ViT-based architectures, such as CrossTeaching \cite{luo2021semi}.
Although the Transformer-based models have demonstrated tremendous success in medical image segmentation, most of them are based on fully-supervised or semi-supervised learning, which generally requires a large amount of fully-annotated training data. To the best of our knowledge, the Transformer-based techniques have not been explored for scribble-supervised medical image segmentation.
\vspace{-4mm}
\subsection{Scribble-supervised Image Segmentation}
To reduce the cost of training a learning model using fully annotated datasets without performance compromise, scribble-supervised learning is widely used in solving various vision tasks, including object detection \cite{Zhang_2020_CVPR, yu2021structure, he2022weakly} and semantic segmentation \cite{Lee_2021_CVPR, Pan_2021_ICCV, wang2023swinmm}. 
Scribble-based supervision has recently emerged as a promising medical image segmentation technique. 
Ji et al. \cite{ji2019scribble} proposed a scribble-based hierarchical weakly supervised learning model for brain tumor segmentation, combining two weak labels for model training, i.e., scribbles on whole tumor and healthy brain tissue, and global labels for the presence of each substructure.
In the meantime, several research works focus on scribble-supervised segmentation without requiring extra annotated masks.  
Can et al. \cite{can2018learning} investigated training strategies to learn parameters of a pixel-wise segmentation network from scribble annotations alone, where a dataset relabeling mechanism was introduced using the dense conditional random field (CRF) during the process of training. 
Luo et al. \cite{luo2022scribble} proposed a scribble-supervised segmentation model via training a dual-branch network with dynamically mixed pseudo labels supervision (DMPLS).
Recently, Cyclemix \cite{Zhang_2022_CycleMix} was proposed for scribble learning-based medical image segmentation, which generated mixed images and regularized the model by cycle consistency.
Generally, none of these methods have exploited global information of the image for the medical image segmentation problem. We believe the hidden global information in the dataset learned by Transformers could be useful for enhancing the performance of segmentation.
\vspace{-2mm}
\section{Method}
\vspace{-1mm}

\subsection{Overview of ScribFormer}
\vspace{-1mm}
A schematic view of the framework of our proposed ScribFormer is presented in Fig. \ref{framework}. 
\textcolor{Revision}{The framework consists of a triple-branch network, i.e., the hybrid CNN and Transformer branches, along with an attention-guided class activation map (ACAM)  branch.}
For scribble-supervised learning, the leveraging dataset $D=\{(x, s)_n\}_{n=1}^N$ consists of images $x$ and scribble annotations $s$, where a scribble contains a set of pixels of strokes representing a certain category or unknown label. \textit{First}, the CNN branch collaborates with the Transformer branch to fuse the local features learned from CNN with the global representations obtained from Transformers, and generates dual segmentation outputs, i.e., $y_{cnn}$ and $y_{Trans}$, which are then compared with the scribble annotations by applying partial cross-entropy loss. \textit{Then}, both outputs are compared with the hard pseudo labels generated by mixing two predictions dynamically for pseudo-supervised learning. \textit{Furthermore}, the process of extracting ACAMs from the CNN branch and verifying the application consisting of ACAMs enables the shallow convolution layer to learn the pixels affected by the deep one. 
Specifically, since the deep convolutional layer can effectively amalgamate the advantages of both CNN and Transformer, it encompasses more advanced local details as well as global contextual information. 
\vspace{-1mm}
\begin{figure}[ht]
\centering 
\setlength{\abovecaptionskip}{-1mm}
\includegraphics[width=\linewidth]{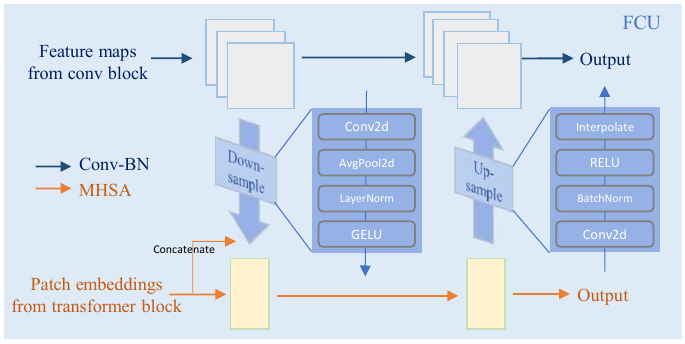}
\caption{\textcolor{Revision}{Schematic illustration of FCU (Feature Coupling Units). Due to inconsistency of feature dimensions of CNN and Transformers, feature maps from convolution blocks and patch embeddings from Transformer blocks are fused in the down-sample and up-sample blocks after the channel and spatial alignments.}}
\label{fcu}
\vspace{-6mm}
\end{figure}
When computing the ACAMs-consistency loss, shallow features are utilized to narrow the gap with the deep features, enabling the shallow features to learn semantic information akin to that present in the deep features. This approach effectively addresses the issue of local activations.

In comparison to previous CNN-Transformer hybrid networks, such as TransUNet \cite{chen2021transunet}, Cotr \cite{xie2021cotr}, and Conformer \cite{peng2021conformer}, our proposed ScribFormer \textit{not only} applies scribble data to the CNN-Trans hybrid network, \textit{but also} takes the unique characteristics of scribble data into account.
Previous networks, such as Conformer \cite{peng2021conformer}, include encoders and decoders as part of the CNN-Trans structure. Our ScribFormer, on the other hand, only integrates the CNN-Trans structure between encoders. In the decoders, the CNN feature and the Transformer feature are distinguished, which allows us to concentrate on similarities between CNN-Trans as a hybrid network while also considering differences in the decoders. This is especially important for the scribble-supervised model lacking supervision signals in scribble annotations (compared to full annotations), which often results in mis-segmentation. Our goal is to ensure both CNN and Transformer branches to focus on different parts of image as much as possible for robust segmentation results.
\vspace{-3mm}
\subsection{Hybrid CNN-Transformer Encoders}

The encoder of the CNN branch adopts a feature pyramid structure. As the stage of the CNN encoder increases, the resolution of the feature map decreases, while the number of channels increases. Each convolution block contains multiple bottlenecks from ResNet, including a down-projection convolution, spatial convolution, and upper-projection convolution. \textcolor{Revision}{The down-projection convolution reduces spatial dimensions of input data by emphasizing crucial information through convolution and max pooling. The spatial convolution extracts features by detecting patterns and correlations among adjacent pixels, enabling the network to capture local features and learn spatial hierarchies. The upper-projection convolution increases the size of feature maps using deconvolution, while preserving spatial relationships of the learned features.} The CNN branch can continuously provide local feature details to the Transformer branch.
Unlike the CNN branch, the Transformer branch concerns global representation, which contains the same number of Transformer blocks as the convolution blocks in the CNN branch. The projection layer compresses the feature map generated by the stem module into patch embeddings. Each Transformer block comprises a multi-head self-attention (MHSA) module and an MLP block, where LayerNorm follows before each layer and also residual connection is used in each layer. 

The FCU (Feature Coupling Units) shown in Fig. \ref{fcu} is introduced to integrate the CNN branch and the Transformer branch for feature fusion. Specifically, the CNN feature map collected from the local convolutional operator and Transformer patch embedding, aggregated with the global self-attention mechanism, is aligned and added.  \textcolor{Revision}{
This alignment ensures that convolutional and Transformer features share the same feature space, preventing issues arising from dimensional disparities. The aligned features are combined through addition, effectively merging locally-captured patterns from the CNN with global contextual relationships from the Transformer. This feature fusion enhances the model's ability to recognize intricate patterns and contextual relationships within the data, achieving effective feature sharing between the two components.} Each Transformer block takes the output of the FCU and adds it to the token embeddings from the previous Transformer block. This process is the same for each CNN block, combining features from dual branches. \textcolor{Revision}{The downsampling process is implemented using Conv2D and AvgPool2D. The Conv features initially traverse a Conv2D layer, followed by an AvgPool2D layer, a layer normalization layer, and a GELU activation layer. Subsequently, they are concatenated with the transformer features from the preceding layer, finalizing the alignment process. Upsampling is executed using both Conv2D and interpolation techniques. Specifically, the transformer features undergo  a sequence of processing, including a Conv2D layer, a batch normalization layer, and a RELU activation layer. The resultant transformer features are then harmonized with the convolutional features via an interpolation operation.}

\begin{figure}[ht]
\centering 
\includegraphics[width=\linewidth]{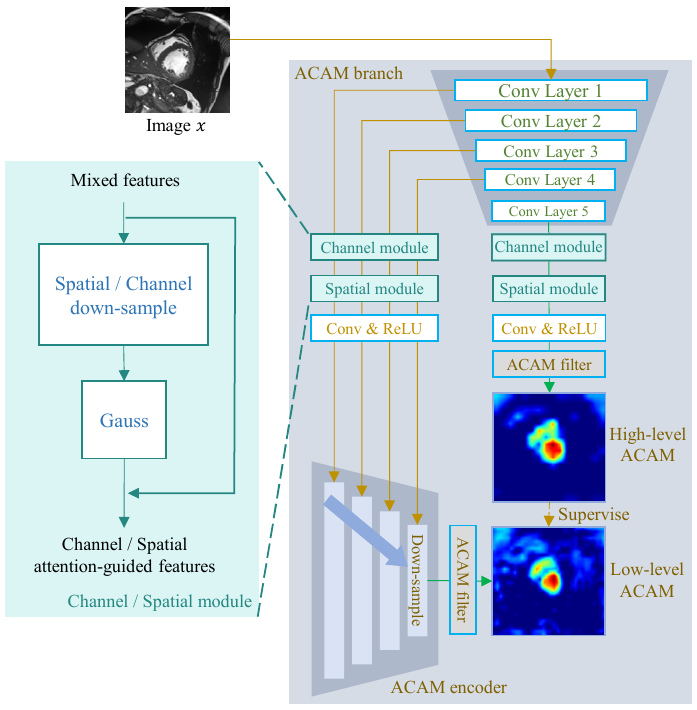}
\caption{\textcolor{Revision}{Schematic illustration of attention-guided class activation maps (ACAM) branch.}}
\label{acam}
\end{figure}

\subsection{Decoders and ACAM Branch}
\subsubsection{Decoders} The structure of the CNN decoder is similar to UNet. The output of each CNN decoder layer is concatenated with the feature map from the last convolutional layer of the corresponding encoder stage. The stem module also contains three convolutional blocks to extract the features required by the decoder. However, unlike UNet decoder, our Transformer decoder upsamples global representation since the \textcolor{Revision}{resolution} of embedding in each Transformer encoder layer is same.

\subsubsection{ACAM Branch} 
As shown in Fig. \ref{acam}, the ACAM branch is designed to identify the most relevant regions on which the training network should concentrate. Compared to traditional CAMs, our attention-guided CAMs are more compatible with semantic segmentation models. \textcolor{Revision}{The images are inputted into Conv Embedding, initiating the process.} The attention-guided CAMs are generated by combining channel attention modulation and spatial attention modulation, which can extract minor features and model the channel-spatial relationship. Specifically, the sensitivity of the features is modeled by the spatial average pooling and the convolutional layer.
\textcolor{Revision}{The Gaussian modulation function in channel attention modulation leverages the distribution of the Gaussian function, which amplifies weights near the mean. This mechanism enhances the importance of regions associated with main features. }
Furthermore, spatial attention modulation is employed to collect spatial interdependency of the features \textcolor{Revision}{through} the channel average pooling and the convolutional layer, which helps increase the minor activations. \textcolor{Revision}{The parameterized representation of the modulation function is :
$f\left ( A\right )=\frac{1}{\sqrt{2 \pi }\sigma }e^{-\frac{\left ( A-\mu \right )^{2}}{2\sigma ^{2}}}$, where attention values $A$ are obtained through spatial/channel down-sampling. }

The attention-guided CAMs (ACAMs) are inspired by attention modulation modules (AMMs) \cite{AMM}, but there are some differences between these two modules. AMMs are connected between convolution stages, while ACAMs are generated for the extra ACAM branch. Moreover, AMMs are generated from local features and optimized for local features, whereas the modulations of ACAMs are generated from the mixture of local features and global representations and are employed to optimize CAMs. \textcolor{Revision}{By incorporating ACAMs, our model leverages strengths of the CNN and Transformer branches  and refines feature localization with a distinctive blend of channel and spatial attention modulation. 
This integration significantly elevates the model's capacity to grasp intricate feature interconnections and extract valuable insights from vital regions, facilitating precise segmentation.}
\vspace{-4mm}
\subsection{Mixed-supervision Learning}
\subsubsection{Scribble-supervised Learning}
 
We apply the partial cross-entropy function for scribble-supervised learning, which ignores unlabeled pixels in the scribble annotation. Hence, the loss of scribble supervision $L_{ss}$ for sample $(x, s)$ is formulated as:
\begin{eqnarray}
\vspace{-1.5mm}
L_{ss}\left(s, y_{CNN}, y_{Trans}\right)=\frac{L_{c e}\left(y_{CNN}, s\right)+ L_{c e}\left(y_{Trans}, s\right)}{2}
\end{eqnarray}
where $y_{CNN}$ is the CNN branch prediction and $y_{Trans}$ is the Transformer branch prediction. $L_{ce}$ is the partial cross-entropy function:
\vspace{-1.5mm}
\begin{eqnarray}
L_{ce}(y, s) & = & \sum_{i \in \Omega_{l}} \sum_{k \in K}-s_{i}^{k} \log \left(y_{i}^{k}\right)
\end{eqnarray}
where $K$ is the set of strategies in scribble annotations, and $\Omega_{l}$ is the set of labeled pixels in scribble $s$. $s_i^k$ and $y_i^k$ are separately the scribble element and the predicted probability of pixel $i$ belonging to class $k$.

\subsubsection{Pseudo-supervised Learning}
Based on difference of receptive fields between the CNN branch and the Transformer branch, we further explore their outputs to boost the model training. Following \cite{luo2022scribble}, the hard pseudo label is generated by dynamically mixing the CNN branch prediction $y_{cnn}$ and the Transformer branch prediction $y_{Trans}$, and then employed to supervise the two predictions separately. The pseudo label loss $L_{pl}$ is formulated as:
\vspace{-1.5mm}
\begin{eqnarray}
L_{pl} = \mathrm{average}\left(L_{dice}\left(y_{CNN}, Y\right), L_{dice}\left(y_{Trans}, Y\right)\right.
\end{eqnarray}
where $L_{dice}$ is the Dice function, and $Y$ is the pseudo label defined by:
\vspace{-1.5mm}
\begin{eqnarray}
Y = \mathrm{argmax}\left(\alpha \times y_{CNN}+\beta \times y_{Trans}\right), \alpha, \beta \in(0,1)
\end{eqnarray}
\textcolor{Revision}{Here, $\alpha$ is dynamically generated using the random.random() function in each iteration, and $\beta$ is set as $1 - \alpha$. By permitting $\alpha$ to vary, the model seeks to discover diverse weight combinations for the branches, with the intention of finding more optimal configurations that reach a balance between the two.}
\begin{figure*}[htbp]
\setlength{\abovecaptionskip}{-1mm}
\centering 
\includegraphics[width=1\textwidth]{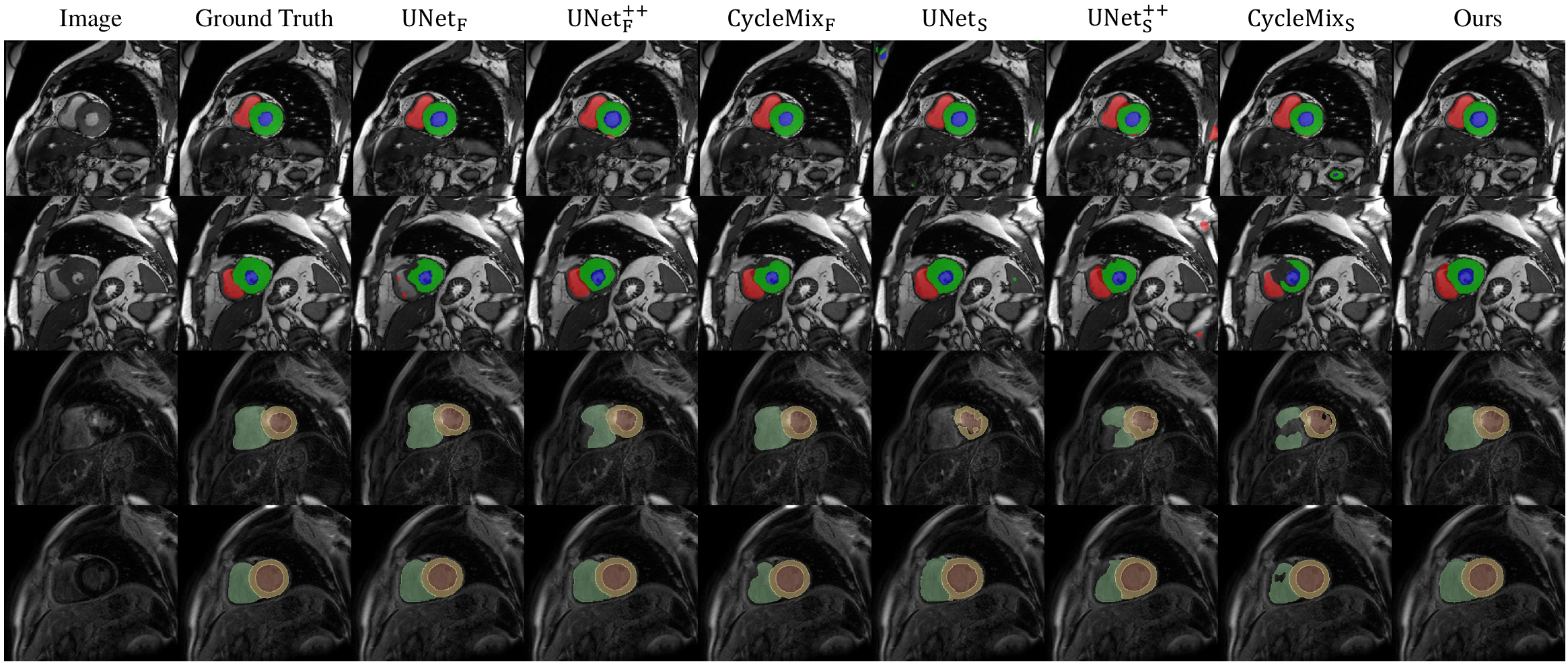}
\caption{Qualitative comparison between ScribFormer and other state-of-the-art methods on ACDC and MSCMRseg datasets. Subscripts \textit{F} and \textit{S} denote the segmentation models trained with fully-annotated masks or scribble annotations, respectively. }
\label{results}
\vspace{-6mm}
\end{figure*}

\subsubsection{ACAM-Consistency Learning}
General consistency learning aims to ensure smooth predictions at a data level, i.e., the predictions of the same data under different transformations and perturbations should be consistent \cite{luo2021semi}. In contrast to data-level consistency, we enforce feature-level consistency through a novel ACAM-consistency evaluation model between the deep features and the shallow features at the pixel level. \textcolor{Revision}{Additionally, this method can also introduce implicit shape constraints.} The ACAM-consistency loss is formulated as:
\vspace{-1.5mm}
\begin{eqnarray}
L_{acam} = \sum_{i} \omega_{i} \times L_{ce}\left(F(E_{i \ldots 4}\left(c_{i}\right), F(c_{5})\right)
\vspace{-3mm}
\end{eqnarray}
It is a weighted sum of a set of cross-entropy losses $L_{ce}$ calculated based on different attention-guided CAMs $c_{i}$ from different layers $i$ of the CNN branch encoder. The convolutional embedding 1 and 2 plus the convolutional layer 1-3 are denoted as $c_1-c_5$, respectively. By aligning different convolutional layers using the ACAM encoder $E$, other ACAMs are expected to be similar to the ACAM of the last convolutional layer $c_5$. The down-sampling layer $i$ of the ACAM encoder is represented by $E_i$, and the number of encoder layers can differ based on the resolutions of ACAMs. The lower the resolution of ACAM, the fewer layers it requires for down-sampling. $E_{i \ldots 4}$ represents that $c_i$ is the input of ACAM encoder layer $i$, and the low-level ACAM is the output from the ACAM encoder layer 4. $F$ is the ACAM filter that is set as the sigmoid function. It should be noted that each pixel of ACAM is labeled either with 1 (if concentrated by the layer) or 0 (not concentrated by the layer).

Finally, the training objective $L_{total}$ is formulated as:
\begin{eqnarray}
L_{total} & = & \lambda_{1} \times L_{ss}+\lambda_{2} \times L_{p l}+\lambda_{3} \times L_{acam}
\end{eqnarray}
where $\lambda_{1}$, $\lambda_{2}$, and $\lambda_{3}$ are the weight factors used to balance different supervisions.
\vspace{-2mm}

\begin{table}[ht!]
  \centering
  \setlength{\abovecaptionskip}{-1mm}
  \caption{Performance Comparison of Dice Score between our Method (ScribFormer) and Other State-of-the-Art Methods on ACDC Dataset. \textcolor{Revision}{Bold denotes the best performance among all methods except nnUNet.}}
  \resizebox{0.88\columnwidth}{!}{%
    \begin{tabular}{l|c|ccc|c}
    \toprule[1pt]
    Method & Data  & LV & RV & MYO  & Avg \\ \hline
    \multicolumn{6}{l}{\textbf{35 scribbles}} \\ \hline
    UNet$_{pce}$ & scribbles & .624 & .537 & .526 & .562 \\
    UNet$_{em}$ & scribbles & .789 & .761 & .788 & .779 \\
    UNet$_{crf}$ & scribbles & .766 & .661 & .590 & .672 \\
    UNet$_{mloss}$ & scribbles & .873 & .812 & .833 & .839 \\
    UNet$_{ustr}$ & scribbles & .605 & .599 & .655 & .620 \\
    UNet$_{wpce}$ & scribbles & .784 & .675 & .563 & .674 \\
    UNet$^{+}_{pce}$ & scribbles & .785 & .725 & .746 & .752 \\
    UNet$^{++}_{pce}$ & scribbles & .846 & .787 & .652 & .761 \\
    Co-mixup & scribbles & .622 & .621 & .702 & .648 \\
    CutMix & scribbles & .641 & .734 & .740 & .705 \\
    Puzzle Mix & scribbles & .663 & .650 & .559 & .624 \\
    Cutout & scribbles & .832 & .754 & .812 & .800 \\
    MixUp & scribbles & .803 & .753 & .767 & .774 \\
    CycleMix$_S$ & scribbles & .883 & .798 & .863 & .848 \\
    \textbf{ScribFormer} & scribbles & \textbf{.922} & \textbf{.871} & .871 & \textbf{.888} \\ \hline
    \multicolumn{6}{l}{\textbf{35 scribbles + 35 unpaired masks}} \\ \hline
    UNet$_D$ & mixed & .404 & .597 & .753 & .585 \\
    MAAG  & mixed & .879 & .817 & .752 & .816 \\
    ACCL  & mixed & .878 & .797 & .735 & .803 \\
    PostDAE & mixed & .806 & .667 & .556 & .676 \\\hline
    \multicolumn{6}{l}{\textbf{35 masks}} \\ \hline
    UNet$_F$ & masks & .892 & .830 & .789 & .837 \\
    UNet$^+_F$ & masks & .849 & .792 & .817 & .820 \\
    UNet$^{++}_F$ & masks & .875 & .798 & .771 & .815 \\
    Puzzle Mix$_F$ & masks &.849 & .807 & .865 & .840 \\
    CycleMix$_F$ & masks & .919 & .858 & \textbf{.882} & .886 \\
    \textcolor{Revision}{nnUNet} &  \textcolor{Revision}{masks} &  \textcolor{Revision}{.943} &  \textcolor{Revision}{.915} &  \textcolor{Revision}{.901} &  \textcolor{Revision}{.920} \\\bottomrule[1pt]
    \end{tabular}%
    }
  \label{ACDCresult}%
  \vspace{-6mm}
\end{table}%

\begin{table}[ht!]
  \centering
  \caption{
Performance Comparison of Dice Score between our Method (ScribFormer) and Other State-of-the-Art Methods on MSCMRseg Dataset. \textcolor{Revision}{Bold denotes the best performance among all methods except nnUNet.}}
  \resizebox{0.88\columnwidth}{!}{%
    \begin{tabular}{l|c|ccc|c}
    \toprule[1pt]
    Method & Data  & LV & RV & MYO  & Avg \\ \hline
    \multicolumn{6}{l}{\textbf{25 scribbles}} \\  \hline
    UNet$^+_{pce}$ & scribbles & .494 & .583 & .057 & .378 \\
    UNet$^{++}_{pce}$ & scribbles & .497 & .506 & .472 & .492 \\
    Co-mixup & scribbles & .356 & .343 & .053 & .251 \\
    CutMix & scribbles & .578 & .622 & .761 & .654 \\
    Puzzle Mix & scribbles & .061 & .634 & .028 & .241 \\
    Cutout & scribbles & .459 & .641 & .697 & .599 \\
    MixUp & scribbles & .610 & .463 & .378 & .484 \\
    CycleMix$_S$ & scribbles & .870 & .739 & .791 & .800 \\
    \textbf{ScribFormer} & scribbles & \textbf{.896} & \textbf{.807} & \textbf{.813} & \textbf{.839} \\  \hline
    \multicolumn{6}{l}{\textbf{25 masks}} \\  \hline
    UNet$_F$ & masks & .850 & .721 & .738 &  .770 \\
    UNet$^+_F$ & masks & .857 & .720 & .689 & .755 \\
    UNet$^{++}_F$ & masks & .866 & .745 & .731 & .774 \\
    Puzzle Mix$_F$ & masks & .867 & .742 & .759 & .789 \\
    CycleMix$_F$ & masks & .864 & .785 & .781 & .810 \\
    \textcolor{Revision}{nnUNet} &  \textcolor{Revision}{masks} &  \textcolor{Revision}{.944} &  \textcolor{Revision}{.880} &  \textcolor{Revision}{.882} &  \textcolor{Revision}{.902} \\\bottomrule[1pt]
    \end{tabular}%
    }
  \label{MSCMRresult}%
  \vspace{-7mm}
\end{table}%

\section{Experiments}
\vspace{-1.5mm}
\subsection{Datasets}
\subsubsection{ACDC}
The ACDC \cite{acdc} dataset consists of cine-MRI scans from 100 patients. For each scan, manual scribble annotations of the left ventricle (LV), right ventricle (RV), and myocardium (MYO) are from \cite{MAAG}. \textcolor{Revision}{The scribble annotations underwent a rigorous process conducted by experienced}
\textcolor{Revision}{annotators.} Following \cite{MAAG, Zhang_2022_CycleMix, shapepu}, the 100 scans are randomly separated into three sets of sizes 70, 15, and 15, respectively, for the purpose of model training, validation and testing. To compare with the state-of-the-art approaches that employ unpaired masks to learn \textcolor{Revision2}{global shape information}, we split the training set into two halves, i.e., 35 training images with scribble labels and 35 masks with full annotations where the corresponding images would not be used for training.
Generally, only 35 training images are used to train the baselines and our ScribFormer unless otherwise specified.
\subsubsection{MSCMRseg}
The MSCMRseg \cite{mscmr1,mscmr2} dataset comprises late gadolinium enhancement (LGE) MRI scans from 45 cardiomyopathy patients. Scribble annotations of LV, MYO, RV for each scan are provided by \cite{Zhang_2022_CycleMix}. The scribble annotations were custom-designed to suit the dataset’s requirements and encompass average coverage percentages 
\begin{table}[ht]
\Large
  \centering
  \caption{
  \textcolor{Revision}{Performance comparison of Dice Score between our method (ScribFormer) and other state-of-the-art methods on HeartUII. Bold denotes the best performance among all methods except nnUNet.}}
  {
    \resizebox{\columnwidth}{!}{%
    \begin{tabular}{l|c|*{6}{c}|c}
    \toprule[1pt]
    Method & Data  & LV & LA & RV & RA & AO & MYO & Avg \\ \hline
    \multicolumn{9}{l}{\textbf{53 scribbles}} \\  \hline
    UNet$_{pce}$ & scribbles & .802 & .833 & .702 & .375 & .694 & .521 & .655   \\
    UNet$_{ustr}$ & scribbles & .709 & .772 & .799 & .389 & .783 & .534 & .664 \\
    UNet$_{em}$ & scribbles & .847 & .865 & .798 & .562 & .814 & .682 & .761 \\
    UNet$_{crf}$ & scribbles & .739 & \textbf{.885} & .812 & .731 & .843 & .698 & .785 \\
    UNet$^{++}_{pce}$ & scribbles & .834 & .819 & .749 & .567 & .694 & .620 & .714\\
    CycleMix$_S$ & scribbles & .851 & .814 & .756 & .799 & \textbf{.871} & .768 & .810   \\
    \textbf{ScribFormer} & scribbles & \textbf{.873} & .867 & \textbf{.859} & \textbf{.774} & .843 & .783 & \textbf{.833} \\
    \hline
    \multicolumn{9}{l}{\textbf{53 masks}} \\  \hline
    UNet$_{F}$ & masks & .771 & .817 & .744 & .714 & .777 & .661 & .747   \\
    UNet$^{++}_{F}$ & masks & .873 & .881 & .825 & .759 & .842 & \textbf{.816} & .833 \\
    nnUNet & masks & .943 & .927 & .886 & .902 & .942 & .882 & .914  
    \\ \bottomrule[1pt]
    \end{tabular}%
    }
}
  \label{HeartUIIresult}%
  \vspace{-6mm}
\end{table}%
for different regions, i.e., background, RV, MYO, and LV scribbles were represented at rates of 3.4\%, 27.7\%, 31.3\%, and 24.1\%, respectively. Compared to ACDC, MSCMRseg is much smaller and more arduous to create, since LGE MRI segmentation is more complicated. Following \cite{Zhang_2022_CycleMix, shapepu}, we randomly divided the 45 scans into three sets: 25 for training, 5 for validation, and 15 for testing.

\subsubsection{HeartUII}
\textcolor{Revision}{HeartUII is a CT dataset collected by us, 
comprising six distinct categories: Right Atrium (RA), Right Ventricle (RV), Left Ventricle (LV), Aorta (AO), Left Atrium (LA), and Myocardium (MYO). To ensure accuracy and authenticity of scribble annotations, we sought the expertise of professionals in the relevant field. These experts utilized ITK-SNAP to meticulously annotate the dataset. The annotation process was  conducted similarly to the ACDC dataset. The dataset consists of a total of 80 cases, with 53 cases utilized for training, 13 for validation, and 16 for testing, respectively. Each case encompasses a range of 78 to 320 slices.}
\vspace{-4mm}
\subsection{Implementation Details}
The model was implemented using Pytorch and trained on one NVIDIA 1080Ti 11GB GPU. \textcolor{Revision}{We initially rescaled the intensity of each slice in the ACDC dataset, the MSCMR dataset, and the HeartUII dataset to a range of values between 0 and 1.} To expand the training set, we applied random rotation, flipping, and noise to the images. The enhanced image was \textcolor{Revision}{adjusted} to 256 $\times$ 256 pixels before being utilized as input to the network. For the MSCMR dataset, each image was cropped or padded to the identical size of 212$\times$ 212 pixels to enhance performance. The optimizer choice was AdamW. \textcolor{Revision}{In a series of preliminary experiments, we observed that the model converged within 300 epochs, with diminishing returns on further training. Therefore, we trained for 300 epochs on each dataset. For learning rate and weight decay, a grid search was conducted, resulting in the optimal performance achieved at a learning rate of 0.001 and weight decay of 0.0005, respectively. Early stopping was not employed during the training process. Additionally, the total training time was hard-coded to maintain consistency across experiments.}
The ACAM-consistency factors $(\omega_{1},\omega_{2},\omega_{3},\omega_{4})$ were set to (0.25, 0.5, 0.75, 1). We empirically set the weights $(\lambda_1, \lambda_2, \lambda_3)$ to (1, 0.5, 0.1) in Eq.(6). For \textcolor{Revision}{all} datasets, Dice Score (Dice) was used as an evaluation metric.
\vspace{-4mm}
\subsection{Comparison with State-of-the-art (SOTA) Methods}
 To demonstrate the comprehensive segmentation performances of our method, the proposed ScribFormer is compared with different SOTA methods. 

We $first$ compared our approach to several state-of-the-art scribble-supervised methods, including 
1) different scribble-supervised training strategies to UNet \cite{unet} as the base segmentation network architecture with only partial cross-entropy loss (pce) \cite{lin2016scribblesup}, using entropy minimization (em) regularization \cite{EM}, with conditional random field (crf) \cite{crf}, with mumford–shah Loss (mloss) \cite{kim2019mumford}, transformation-consistent regularization (ustr) \cite{liu2022weakly}, 
and weighted partial cross-entropy loss (wpce) \cite{MAAG}, 
or utilizing uncertainty-aware self-ensembling; 
2) different frameworks with same scribble-supervised training loss, i.e., using partial cross-entropy loss on different variants of UNet$_{pce}$ \cite{lin2016scribblesup},  including UNet$^{+}_{pce}$ \cite{unet+}, which has fewer channels in the upsampling path with transpose convolutions adjusted to match the number of classes, and UNet$^{++}_{pce}$ \cite{unet++}, a classic variant incorporating nested and dense skip connections upon original UNet architecture;
3) different data augmentation strategies to UNet$^{+}_{pce}$ \cite{unet+} , including Co-mixup \cite{comixup}, CutMix \cite{cutmix}, Puzzle Mix \cite{puzzlemix}, Cutout \cite{cutout}, MixUp \cite{mixup}, or CycleMix$_S$ \cite{Zhang_2022_CycleMix}.
$Second$, we also compared our method with some adversarial learning methods, including UNet$_D$ \cite{MAAG}, MAAG \cite{MAAG}, ACCL \cite{accl}, and PostDAE \cite{postdae}, which utilized additional unpaired segmentation masks.
$Finally$, we investigated the upper bound using all mask annotations, i.e., fully-supervised methods such as UNet$_F$ \cite{unet}, UNet$^+_F$ \cite{unet+},  UNet$^{++}_F$ \cite{unet++}, and those applying augmentation strategies such as Puzzle Mix$_F$ \cite{puzzlemix} and CycleMix$_F$ \cite{Zhang_2022_CycleMix}. 

The results of the above methods on ACDC and MSCMR are reported in Table~\ref{ACDCresult}, Table~\ref{MSCMRresult} \textcolor{Revision}{and Table~\ref{HeartUIIresult}} separately, with some results obtained from \cite{luo2022scribble} and \cite{Zhang_2022_CycleMix}. 
In the initial section of \textcolor{Revision}{these three tables}, our ScribFormer model showcases its superiority over several training strategies, model architectures, and data augmentation techniques based on UNet when it comes to scribble supervision. Notably, it outperforms the state-of-the-art method, CycleMix, by a substantial margin of 4.0\% (88.8\% vs 84.8\%), 3.9\% (83.9\% vs 80.0\%), \textcolor{Revision}{and 2.3\% (83.3\% vs 81.0\%) } on ACDC, MSCMRseg, \textcolor{Revision}{and HeartUII}, respectively. This compelling performance differential underscores the effectiveness of incorporating Transformer global context into CNN's local features within the framework of scribble-supervised semantic segmentation.

In the second section of Table~\ref{ACDCresult}, the ACDC results underscore substantial performance advancements achieved by ScribFormer compared to other weakly-supervised methods. Notably, ScribFormer's Dice scores for all three categories (LV, MYO, and RV) outperform the previous best method (MAAG). Unlike approaches relying on additional unpaired masks, which are constrained in learning \textcolor{Revision2}{global shape information} from a limited training image set, ScribFormer overcomes this limitation. It achieves this by leveraging the ACAM branch to implicitly learn \textcolor{Revision2}{global shape information}, eliminating the need for extra fully-annotated masks.

In the final sections of all three tables, we conducted comparison between ScribFormer and several fully-supervised learning methods, including CycleMix \textcolor{Revision}{and nnUNet} under full supervision. As observed in the tables, fully-supervised learning outperforms scribble annotations combined with additional unpaired masks. This performance difference is primarily attributed to the exclusion of images associated with masks and the absence of pixel-wise relationships.
\textcolor{Revision}{However, it's worth noting that our ScribFormer outperforms most of the fully-supervised methods (except nnUNet) at a lower annotation cost. This demonstrates the great potential of the proposed scribble-supervised model in medical image segmentation.}

Fig.~\ref{results} presents segmentation results of different methods on ACDC and MSCMR. It can be observed that other scribble-supervised methods tend to generate insufficient or extra segmentation areas, especially on MSCMR, probably due to the limited image information learned from scribbles. In contrast, our method can obtain global representations from the Transformer branch, making up for the deficiency of CNN local features. The results generated by our method are closer to the ground truth, especially in terms of shape completeness than other scribble-supervised and even fully-supervised methods.

\vspace{-3mm}
\subsection{Comparison with Pseudo-label Generating Methods}

\subsubsection{Comparison with UNet-based Methods}
To assess the performance of ScribFormer in comparison to other methods for pseudo-label generation, we adopted a UNet with only partial cross-entropy loss (pce) \cite{lin2016scribblesup} as the foundation, enhanced in several ways: 1) UNet$_{rw}$ \cite{grady2006random}, utilizing pseudo-labels generated by the Random Walker method. 2) UNet$_{s2l}$ \cite{S2L}, incorporating pseudo-labeling alongside label filtering known as Scribble2Label. 3) DMPLS \cite{luo2022scribble}, employing a dual-branch approach with dynamically mixed pseudo-label supervision. 4) TS-UNet \cite{can2018learning}, a variant of UNet++ that combines the Random Walker, Dense CRF, and uncertainty estimation techniques.
Table~\ref{unet} presents the results. It's evident that some pseudo-label-based methods using scribble annotations can achieve reasonably good performance, with both S2L and DMPLS achieving accuracy of 80\% or higher. However, our approach outperforms CNN-based methods by a substantial margin, underscoring the effectiveness of the CNN-Transformer synergy embedded in our network.

\vspace{-4mm}
\begin{table}[!htbp]
 \LARGE
  \centering
    \caption{
Comparison of dice score with pseudo-label Generating Methods on the ACDC Dataset.}
  \resizebox{0.9\columnwidth}{!}{%
    \begin{tabular}{l|c|c|ccc|c}
    \toprule[1pt]
    Method & Backbone &Data  & LV & RV & MYO & Avg \\
    \midrule
    UNet$_{pce}$ & CNN &scribbles & .624 & .537 & .526 & .562 \\
    UNet$_{rw}$ & CNN & scribbles & .840 & .730 & .802 & .791 \\
    UNet$_{s2l}$ & CNN & scribbles & .767 & .715 & .765 & .820 \\
    DMPLS & CNN & scribbles & .875 & \textbf{.903} & .852 & .870 \\
    TS-UNet & CNN & scribbles & .479 & .408 & .272 & .386 \\
    \hline
    \textcolor{Revision}{SwinUNet} & \textcolor{Revision}{Trans} & \textcolor{Revision}{scribbles} & \textcolor{Revision}{.872} & \textcolor{Revision}{.773} & \textcolor{Revision}{.793} & \textcolor{Revision}{.813} \\ \hline
    \textcolor{Revision}{TransUNet} & \textcolor{Revision}{CNN+Trans} & \textcolor{Revision}{scribbles} & \textcolor{Revision}{.857} & \textcolor{Revision}{.762} & \textcolor{Revision}{.807} & \textcolor{Revision}{.808} \\ 
    \textcolor{Revision}{TFCNs} & \textcolor{Revision}{CNN+Trans} & \textcolor{Revision}{scribbles} & \textcolor{Revision}{.839} & \textcolor{Revision}{.713} & \textcolor{Revision}{.774} & \textcolor{Revision}{.775} \\
    \textbf{ScribFormer} & CNN+Trans & scribbles & \textbf{.922} & .871 & \textbf{.871} & \textbf{.888} \\
    \bottomrule
    \end{tabular}%
    }
  \label{unet}%
  \vspace{-3mm}
\end{table}%
\subsubsection{Comparison with Transformer-based Methods}

In this section, we further compared our method with Transformer-based methods in scribble-annotated medical image. Specifically, 
SwinUNet \cite{swinunet} are the volumetric medical image segmentation models utilizing pure Transformers as the encoder to capture long-range spatial dependencies. Meanwhile, TransUNet \cite{chen2021transunet} and TFCNs \cite{li2022tfcns} are both planar medical image segmentation models utilizing a combination of convolutional layers and Transformers.
\textcolor{Revision}{For fairness, all models were trained using the labeled pixels from the scribble data and incorporated pseudo labels generated by the Random Walker algorithm.}
Table~\ref{unet} contains the outcome of our experiments. Interestingly, the Transformer-based medical image segmentation models, which were designed with full annotation data in mind, demonstrated only average performance when applied to scribble data. In contrast, our ScribFormer model excelled in this context, achieving superior performance by adeptly combining both local detailed information and global contextual understanding.

\vspace{-3mm}
\begin{figure*}[htbp]
\setlength{\abovecaptionskip}{-1mm}
\centering 
\includegraphics[width=\textwidth]{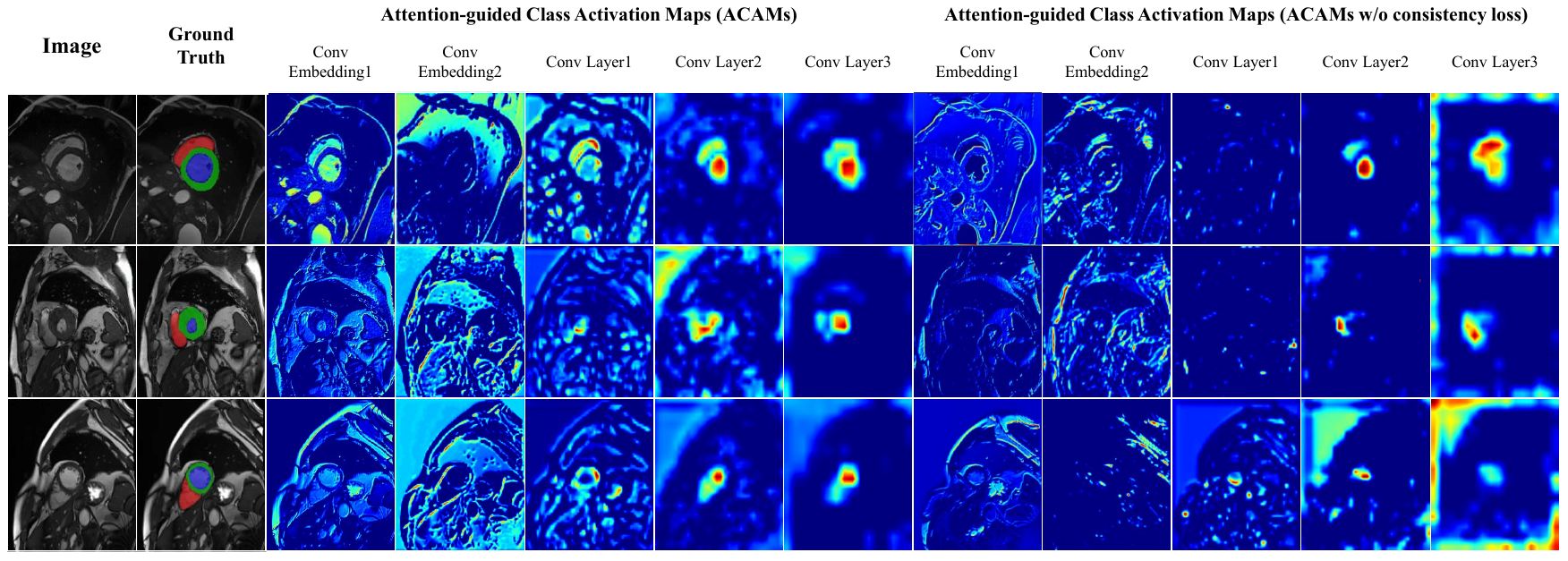}
\caption{\textcolor{Revision}{The comparison of Attention-guided Class Activation Maps (ACAMs) between different layers of ScribFormer on the ACDC dataset.}}
\label{cam}
\vspace{-5mm}
\end{figure*}
\subsection{Ablation Study}
This section studies the effectiveness of different components of the proposed ScribFormer, including the CNN, Transformer, and ACAM branches. Table~\ref{ablationl} reports the results.
\subsubsection{Effectiveness of Transformer Branch} 
As illustrated in Table~\ref{ablationl}, Model \#4 exhibits a significantly better performance than Model \#1 \textcolor{Revision}{and Model \#2}. 
For Model \#1, it is difficult to obtain global representations from scribble annotations by using CNN. 
\textcolor{Revision}{For Model \#2, the pure Transformer architecture excels in capturing global information, granting it a distinct advantage when dealing with irregular regions such as MYO during segmentation.} On the other hand, the CNN branch of Model \#4 provides local features to minimize incorrect predictions of unlabeled pixels within the object. Meanwhile, Transformer branch of Model \#4 provides global representations that help reduce incorrect predictions of unlabeled pixels throughout the entire image, including the background.
\vspace{-3mm}
\begin{table}[ht!]
\LARGE
  \centering
  \setlength{\abovecaptionskip}{0mm}  
  \setlength{\belowcaptionskip}{-2mm}
  \caption{\textcolor{Revision}{Ablation Study of ScribFormer for Image Segmentation Using Dice Score, Investigating Different Settings including the CNN Branch, Transformer Branch, and ACAM Branch.}}
  \resizebox{0.9\columnwidth}{!}{%
    \begin{tabular}{c|ccc|ccc|c}
    \toprule[1pt]
   Models & CNN  & Transformer & ACAM & LV & RV & MYO & Avg \\
   \midrule
    \#1   & \checkmark     & $\times$     & $\times$   & .809 & .642 & .582 & .678 \\
    \textcolor{Revision}{\#2}   & \textcolor{Revision}{$\times$}     & \textcolor{Revision}{\checkmark}     & \textcolor{Revision}{$\times$}     & \textcolor{Revision}{.790} & \textcolor{Revision}{.701} & \textcolor{Revision}{.525} & \textcolor{Revision}{.672} \\
    \#3   & \checkmark     & $\times$      & \checkmark     & .830 & .659 & .650 & .713 \\
    \#4   & \checkmark    & \checkmark     & $\times$      & .906 & .862 & .847 & .872 \\
    \#5   & \checkmark     & \checkmark     & \checkmark     & .922 & .871 & .871 & .888 \\\bottomrule[1pt]
  \end{tabular}%
  }
  \label{ablationl}%
\end{table}%
\vspace{-3mm}
\subsubsection{Effectiveness of ACAM Branch}
As shown in Table~\ref{ablationl}, compared to Model \#1, Model \#3 with the extra ACAM branch achieves better results. The same situation occurred between Model \#4 and Model \#5. Since the unlabeled pixels in the scribble do not participate in the training, it is difficult for the model to predict these pixels. On the contrary, ACAM can obtain the pixels with more attention by the convolution layer, which can expand the trainable pixels to the entire image. In addition, the proposed ACAM-consistency loss can train the low-level convolutional layers under the supervision of high-level convolutional features, leading to further improvement in model performance.

\subsubsection{Effectiveness of Decoder}
\textcolor{Revision}{
As depicted in Table \ref{decoder_ablation}, we conducted ablation experiments involving different decoding strategies built upon the foundation of the CNN-Transformer encoder. Specifically, we assessed the performance when} \textcolor{Revision}{ employing only CNN as the decoder, solely Transformer as the decoder, and a combination of both CNN and Transformer as decoders. The results unequivocally affirm the effectiveness of our multi-branch decoder design in enhancing segmentation performance. Notably, the CNN-Transformer hybrid decoderoutperforms both individual decoders, substantiating the claim made in the second paragraph of Section III-A. In that section, we emphasize the hybrid design's ability to focus on the shared aspects between the CNN and Transformer components while accommodating the unique characteristics of each decoder. This design consideration proves particularly vital in the context of scribble-supervised models, where robustness against mis-segmentation is achieved through tailored attention to various parts of the image. These results reinforce the significance of our approach in achieving superior segmentation accuracy.
}
\vspace{-4mm}
\begin{table}[ht!]
  \small
  \centering
  \caption{
\textcolor{Revision}{Comparison of Performance with CNN Decoder and Transformer Decoder on the ACDC Dataset Using Dice Score.}}
  \resizebox{0.93\columnwidth}{!}{%
    \begin{tabular}{l|c|ccc|c}
    \toprule[1pt]
    Decoder & Data  & LV & RV & MYO & Avg \\
    \midrule
    CNN & scribbles & .748  & .654  & .675  & .692 \\
    Transformer & scribbles & .869 & .804 & .818 & .830 \\
    CNN+Transformer & scribbles & .922 & .871  & .871 & .888 \\
    \bottomrule[1pt]
    \end{tabular}%
    }
  \label{decoder_ablation}%
  \vspace{-2mm}
\end{table}%

\subsubsection{Effectiveness of Loss Function}
\textcolor{Revision}{As shown in Table \ref{ablation_loss}, to comprehensively examine the effects of various loss functions on the overall performance of our model, we systematically assess the influence of each loss function on the Dice score. Our investigations provide insights into the role of each loss function in enhancing the model's stability and overall segmentation accuracy. Notably, the incorporation of the pseudo-label loss ($L_{pl}$) leads to the most substantial performance improvement, resulting in a notable 8.6\% enhancement compared to methods solely utilizing the loss ($L_{ss}$). Furthermore, the inclusion of the $L_{acam}$ loss helps mitigate the performance discrepancy across different categories.}
\vspace{-5.5mm}
\begin{table}[ht!]
  \centering
  \caption{\textcolor{Revision}{Ablation study on the loss function using Dice Score.}}
    \resizebox{0.88\columnwidth}{!}{
    \begin{tabular}{ccc|ccc|c}
    \toprule[1pt]
    $L_{ss}$  & $L_{pl}$ & $L_{acam}$ & LV & RV & MYO & Avg \\
   \midrule
     \checkmark     & $\times$     & $\times$   & .822 & .747 & .771 & .780 \\
    \checkmark    &  $\times$  & \checkmark     & .786 & .801 & .831 & .806 \\
    \checkmark     & \checkmark   &  $\times$    & .907 & .854 & .837 & .866 \\
    
     \checkmark    & \checkmark     & \checkmark     & .922 & .871 & .871 & .888 \\\bottomrule[1pt]
  \end{tabular}
  }
  \label{ablation_loss}%
  \vspace{-4mm}
\end{table}%

\subsubsection{Effectiveness of $\lambda$ and $\omega$ }
\textcolor{Revision}{To investigate the influence of $\lambda$ and  $\omega$ values on model performance, we carried out a series of ablation experiments targeting these parameters. Beginning with $\lambda$, it's important to note that $\lambda_1$ should be no greater than 1. To explore its impact, we reduced $\lambda_1$ to 0.9 while adjusting $\lambda_2$ to 0.3. The findings, as presented in Table \ref{ablationl_lambda}, indicate that decreasing $\lambda_1$ and $\lambda_2$ results in decreased performance. This observation emphasizes the advantage of setting $\lambda_1$ and $\lambda_2$ to higher values for better performance. As for $\omega$ values, which should follow an arithmetic progression within the range [0, 1], we specifically reduced $w_4$ to 0.9. We then reconfigured the arithmetic progression as $(\omega_1, \omega_2, \omega_3, \omega_4) = (0.225, 0.45, 0.675, 0.9)$ and conducted corresponding experiments. The results indicated a performance decline, as seen in Table \ref{ablationl_omega}, upon altering $\omega_4$ to smaller one.} \textcolor{Revision2}{Additionally, significance tests were conducted, revealing that the obtained p-values for both experiments were greater than 0.05. This may be attributed to the influence of extremely small values and limited sample size in the experimental data. We acknowledge this potential impact in our method.}

\vspace{-5mm}

\begin{table}[ht!]
  \centering
  \caption{\textcolor{Revision}{Ablation study on the $\lambda$ using Dice Score.}}
    \resizebox{0.78\columnwidth}{!}{
    \begin{tabular}{ccc|ccc|c}
    \toprule[1pt]
    $\lambda_{1}$  & $\lambda_{2}$ & $\lambda_{3}$ & LV & RV & MYO & Avg \\
   \midrule
     1     &    0.5      & 0.1   & .922  & .871 & .871 & \textbf{.888} \\
     0.9     &  0.3      & 0.1   & .917  & .866 & .871 & .885  \\
   \bottomrule[1pt]
  \end{tabular}
  }
  \label{ablationl_lambda}%
  \vspace{-3mm}
\end{table}%
\vspace{-5mm}
\begin{table}[ht!]
  \centering
  \caption{\textcolor{Revision}{Ablation study on the $\omega$ using Dice Score.}}
  \resizebox{0.94\columnwidth}{!}{
    \begin{tabular}{cccc|ccc|c}
    \toprule[1pt]
    $\omega_{1}$ &$\omega_{2}$  & $\omega_{3}$  & $\omega_{4}$  & LV & RV & MYO  & Avg \\
   \midrule
     0.25   & 0.5  &  0.75    & 1   & .922  & .871 & .871 & \textbf{.888}  \\
     0.225  & 0.45 &  0.675   & 0.9    & .921  & .870 & .868 & .886 \\
   \bottomrule[1pt]
  \end{tabular}
  }
  \label{ablationl_omega}%
  \vspace{-2mm}
\end{table}
\vspace{-6mm}
\subsection{ACAMs Visualization}
\color{black}{To explain the role of ACAM-consistency and further verify the effectiveness of Transformers, the visualization of the ACAMs in each layer is shown in Fig. \ref{cam}. It can be observed that i) the ACAMs of} convolution layer3 closely match the goal segmentation region of the ground truth, rather than discriminative regions, which means the introduction of  Transformers can help modulate the activation maps, emphasizing global features in scribble supervision. ii) As the network goes deeper, the activation maps of the convolution layer also gradually approach the target segmentation areas. Specifically, \textcolor{Revision}{Conv Embedding1 and Conv Embedding2} concentrate on locating high-contrast regions, which appear as low and scattered highlights on the activation map. The activation maps of the Conv Layer1 contain multiple relatively-dense tiny regions and begin to focus on the segmentation area. Conv Layer2 gets closer to the target, and the ACAMs of Conv Layer3 are extremely similar to the ground truth. 
The observed outcome can be ascribed to the joint effect of Transformer refinement and ACAM-consistency regularization on the attention regions of the shallow ACAMs. \textcolor{Revision}{Furthermore, when comparing ACAM with and without consistency loss, it is evident that our model  maintains the capability to focus on the target region even without consistency loss. Nonetheless, certain level of confusion arises in the absence of consistency loss. This highlights the effectiveness of integrating our ACAMs with consistency loss, as it serves to further enhance the refinement of attention-guided class activation maps.}
\vspace{-4mm}
\subsection{Data Sensitivity Study}
The data sensitivity study delves into ScribFormer's performance when trained with varying numbers of scribble annotations. Table~\ref{sensitivity} showcases a clear trend where ScribFormer's performance progressively improves as the number of scribble-annotated samples increases. Notably, even with just 14 training samples that include scribbles, our model achieves a respectable accuracy of 84.7\%. This highlights ScribFormer's ability to produce satisfactory segmentation results with a relatively small amount of scribble annotations. The model's overall performance stabilizes as it's trained with 56 scribble annotations (which amounts to 80\% of the total 70 scribbles). The peak performance is achieved when all 70 scribble annotations are utilized, resulting in an impressive accuracy of 89.4\%.
\vspace{-5mm}
\begin{table}[htbp]
  \centering
  \caption{Data Sensitivity Study: Evaluating ScribFormer's Performance with Varying Numbers of Scribbles for Training Using Dice Score.}
  \resizebox{0.72\columnwidth}{!}{%
  {%
    \begin{tabular}{c|ccc|c}
    \toprule[1pt]
    Data  & LV & RV & MYO & Avg \\
    \midrule
    \multicolumn{1}{c|}{14 scribbles} & .899  & .839  & \multicolumn{1}{c|}{.804}  & .847 \\
    \multicolumn{1}{c|}{28 scribbles} & .900  & .853  & \multicolumn{1}{c|}{.844}  & .866 \\ 
    \multicolumn{1}{c|}{35 scribbles} & .922  & .871  & \multicolumn{1}{c|}{.871}  & .888 \\
    \multicolumn{1}{c|}{56 scribbles} & .925  & .873  & \multicolumn{1}{c|}{.877}  & .892 \\
    \multicolumn{1}{c|}{70 scribbles} & .926  & .878  & \multicolumn{1}{c|}{.877}  & .894 \\
    \bottomrule[1pt]
    \end{tabular}%
    }}
  \label{sensitivity}%
  \vspace{-4mm}
\end{table}%

\vspace{-3mm}
\subsection{Model Complexity Comparison}

\textcolor{Revision}{As illustrated in Table \ref{complex}, to assess the model's complexity, we compared the parameter count and FLOPs between the proposed ScribFormer and other benchmark methods. It's worth noting that the UNet variants, such as UNet$_{pce}$, UNet$_{ustr}$, and UNet$^{++}_{pce}$, maintain equivalent parameter sizes and FLOPs to their respective UNet and UNet$^{++}$ counterparts.
Compared with the UNet variants, the parameter count of our model is relatively higher, primarily due to the inclusion of Transformer components. However, in comparison to CycleMix, our model exhibits lower computational complexity. 
Furthermore, we evaluated the averaged inference time per case within the HeartUII test set for both CycleMix and ScribFormer. The results indicate that CycleMix requires 21.21 seconds per case, whereas ScribFormer achieves a faster inference time at just 13.96 seconds. The observation underscores our advantage in terms of inference efficiency. \textcolor{Revision2}{And, we observe computational demands of the Transformer architecture posing a potential challenge for real-time applications. To address this concern, our ongoing efforts are focusing on  optimization of ScribFormer to enhance its suitability across a broader spectrum of scenarios. At the same time, experimental results also suggest that ScribFormer outperforms or competes favorably with existing architectures in some benchmark tasks. These evidences add credibility to the model's capabilities, reinforcing its potential as a reliable solution in various applications.}
}%
\vspace{-4mm}
\begin{table}[ht!]
  \centering
  \caption{
\textcolor{Revision}{Model Complexity Comparison between Our Method (ScribFormer) and other methods On the
HeartUII Dataset.}}
  \resizebox{0.6\columnwidth}{!}{%
    \begin{tabular}{l|ccc}
    \toprule[1pt]
    Method & Params(M) & Flops(G)  \\
    \midrule
    \textcolor{Revision}{UNet} &  1.81 & 24.25   \\
    \textcolor{Revision}{UNet$^{++}$} & 9.16 & 279.25\\
    \textcolor{Revision}{CycleMix} & 25.76 & 469.41 \\
    \textcolor{Revision}{ScribFormer} & 50.44 &436.67  \\
    \bottomrule[1pt]
    \end{tabular}%
    }
  \label{complex}%
\end{table}
\vspace{-6mm}
\begin{table}[ht!]
\small
  \centering
  \caption{
\textcolor{Revision2}{Comparison of Confidence Intervals (CI) and p-values between Our Method (ScribFormer) and Other Methods on the HeartUII Dataset. The p-values were obtained by conducting t-tests between ScribFormer and other methods. Therefore, the p-value for ScribFormer is null.}}
    \begin{tabular}{l|cc}
    \toprule[0.6pt]
    Method & Dice (95\% CI) & p-value \\
    \midrule[0.4pt]
    \textcolor{Revision}{UNet$_{pce}$} &  .655 (.609 to .694) & \textcolor{Revision3}{5.380 $\bigcdot$ 10$^{-9}$}\\
   \textcolor{Revision}{UNet$_{ustr}$} & .664 (.621 to .703)  & \textcolor{Revision3}{9.430 $\bigcdot$ 10$^{-9}$} \\
    \textcolor{Revision}{UNet$_{em}$} & .761 (.729 to .793) & \textcolor{Revision3}{4.026 $\bigcdot$ 10$^{-4}$}\\
    \textcolor{Revision}{UNet$_{crf}$} & .785 (.720 to .839) & \textcolor{Revision3}{1.080 $\bigcdot$ 10$^{-1}$}\\
    UNet$^{++}$$_{pce}$ & .714 (.670 to .757) & \textcolor{Revision3}{1.064 $\bigcdot$ 10$^{-5}$} \\
    \textcolor{Revision}{CycleMix$_{S}$} & .810 (.790 to .831) & \textcolor{Revision3}{1.073 $\bigcdot$ 10$^{-1}$}\\
   \textcolor{Revision}{ScribFormer} & .833 (.808 to .854)  & / \\
    \bottomrule[0.6pt]
    \end{tabular}
  \label{CI}%
  \vspace{-2mm}
\end{table}%
\vspace{-5mm}
\subsection{Inference Statistical Evaluation}
\begin{figure}[htbp]
\centering
\includegraphics[width=\linewidth]{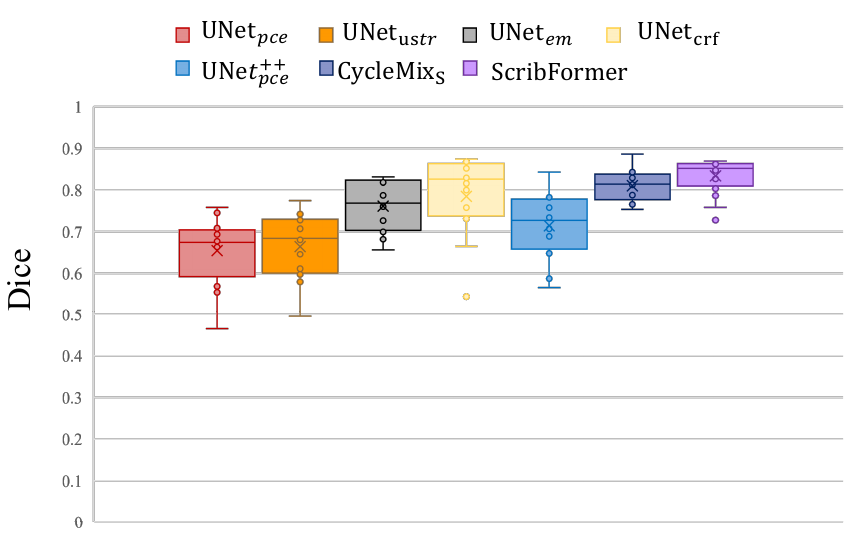}
\caption{\textcolor{Revision}{Comparison of the inference results for each case in the test set with other SOTA methods on the HeartUII dataset.}}
\label{bootstrap}
\vspace{-6mm}
\end{figure}

\textcolor{Revision3}{To conduct a thorough significance analysis, we computed 95\% confidence intervals using the bootstrap method \cite{efron1987better} and calculated p-values through t-test on the HeartUII testing set, as presented in Table \ref{CI}.} By comparing 95\% confidence intervals and p-values, our approach exhibits significant differences compared to UNet$_{pce}$, UNet$_{ustr}$, UNet$_{em}$, and UNet$_{pce}^{++}$.  
Despite the non-significant trend in p-values for UNet$_{crf}$ and CycleMix$_{S}$, analyzing the 95\% confidence interval reveals a narrower range for our method compared to UNet$_{crf}$. This indicates lower overall variance and also suggests greater robustness in our model. Moreover, by examining the box plot of inference results in Fig. \ref{bootstrap}, our method demonstrates a higher median than CycleMix$_{S}$, indicating that our approach outperforms CycleMix$_{S}$ at the average level of the testing samples.

\vspace{-2mm}
\section{Conclusion}
In this paper, a new Transformer-CNN hybrid solution, called ScribFormer, has been proposed to solve the limitations of CNN-based networks for scribble-supervised medical image segmentation.
The main motivation behind ScribFormer is based on our observation that attention weights from shallow Transformer blocks could capture low-level spatial feature similarities, while attention weights from deep Transformer blocks could capture high-level semantic context. 
Specifically, ScribFormer explicitly leverages the attention weights from the Transformer branch to refine both the convolutional features and the ACAMs generated from the CNN branch. 
\textcolor{Revision}{Our method, as the first Transformer-based solution in scribble-supervised medical image segmentation, is simple, efficient, and effective for generating high-quality pixel-level segmentation results. \textcolor{Revision}{It enhances medical image analysis by reducing the need for extensive annotations, thereby minimizing manual labeling efforts and broadening the possibilities for scribble-supervised medical image segmentation.} Experimental results demonstrate new SOTA performance of our ScribFormer on ACDC, MSCMRseg\textcolor{Revision}{, and HeartUII datasets.} \textcolor{Revision2}{However, it should be noted that our method may yield non-significant results when compared with some SOTA methods for inference statistical evaluation. In future work, we will focus on addressing limitations of our method by further reducing its computational complexity and exploring the influence of hyperparameters more comprehensively.}}

\bibliographystyle{IEEEtran}
\bibliography{main}

\end{document}